\documentclass{article}

\usepackage[dblblindworkshop,final]{neurips_2025}

\workshoptitle{MATH-AI}

\usepackage{natbib}
\usepackage[utf8]{inputenc} % allow utf-8 input
\usepackage[T1]{fontenc}    % use 8-bit T1 fonts
\usepackage{hyperref}       % hyperlinks
\usepackage{url}            % simple URL typesetting
\usepackage{multirow}       % for multirow cells in tables
\usepackage{titletoc}

\usepackage{booktabs}       % professional-quality tables
\usepackage{amsfonts}       % blackboard math symbols
\usepackage{nicefrac}       % compact symbols for 1/2, etc.
\usepackage{microtype}      % microtypography
\usepackage{xcolor}         % colors
\usepackage{comment}
\usepackage{graphicx}
\usepackage{svg}
\usepackage{float}
\usepackage{placeins}
\usepackage{algorithm}
\usepackage{algorithmicx}
\usepackage{algpseudocode}
\usepackage{amsmath}
\usepackage{listings}
\newcommand{\nope}[1]{}
\newcommand{\approachname}{DiagramIR}

\title{\approachname: An Automatic Pipeline for Educational Math Diagram Evaluation}

\author{Vishal Kumar$^{1, }$\thanks{Corresponding author: \texttt{vishalku@stanford.edu}. All code associated with this project can be found \href{https://github.com/EduNLP/DiagramIR}{here}.}, Shubhra Mishra$^{2}$, Rebecca Hao$^{1}$, Rizwaan Malik$^{1}$, David Broman$^{2}$, \\ \textbf{Dorottya Demszky$^{1}$}\\
Stanford University$^1$, KTH Royal Institute of Technology$^2$}

\begin{document}

\maketitle
\begin{abstract}
  Large Language Models (LLMs) are increasingly being adopted as tools for learning; however, most tools remain text-only, limiting their usefulness for domains where visualizations are essential, such as mathematics. Recent work shows that LLMs are capable of generating code that compiles to educational figures, but a major bottleneck remains: scalable evaluation of these diagrams. We address this by proposing \emph{\approachname}: an automatic and scalable evaluation pipeline for geometric figures. Our method relies on intermediate representations (IRs) of LaTeX TikZ code. We compare our pipeline to other evaluation baselines such as LLM-as-a-Judge, showing that our approach has higher agreement with human raters. This evaluation approach also enables smaller models like GPT-4.1-Mini to perform comparably to larger models such as GPT-5 at a 10x lower inference cost, which is important for deploying accessible and scalable education technologies.
\end{abstract}

\vspace{-0.5em}
\section{Introduction}
\vspace{-0.5em}
Large Language Models (LLMs) have garnered significant attention for their potential to enhance education \cite{kasneci2023chatgpt, labadze2023role}, and recent studies have shown that students are increasingly adopting AI as part of their learning process \cite{zhou2024teachers, lee2024cheating}. Despite this, the current wave of LLM-assisted learning tools remains narrow: most rely on chatbot-style interfaces, where text is the sole input and output. While this paradigm has been useful for non-visual subjects, it can be limiting for highly visual subjects like mathematics, where diagrams and spatial reasoning play a key role in students' understanding of problems and concepts \cite{hegarty1999types}. 

Recent work has examined how LLMs can be used to generate pedagogically-useful mathematical diagrams \cite{malik2025mathematikz, lee2025text}, but they face a critical challenge: evaluation. Evaluating whether a mathematical figure is \emph{useful} is challenging: it involves (1) evaluating whether the diagram aligns with the prompt, (2) whether the diagram is mathematically and visually sound, and (3) whether the diagram is pedagogically meaningful given the context. The first and third aspects are difficult given their subjective nature and the complexity of pedagogical evaluation, requiring further study. We focus in this paper on the second aspect: scalable evaluation of whether a diagram is mathematically and visually sound. Current techniques largely rely on human evaluation \cite{malik2025mathematikz}, which provide high quality insights for static projects, but cannot be deployed scalably in live chatbot-like application settings.

One promising direction in these settings has been to apply LLM-as-a-Judge methods, which have proven effective in text-based evaluation tasks \cite{zheng2023judging}. However, extending this paradigm to diagrams introduces unique obstacles. Unlike text, diagrams require reasoning over geometry, proportions, and spatial layout, which are all features that LLMs cannot directly process without translation into symbolic or textual representations. Multimodal Language Models (MLMs)\citep{chen2024mllm} are a natural alternative, but benchmarks like MathVista \cite{lu2023mathvista} and MathVerse \cite{zhangMathVerseDoesYour2024} demonstrate that current MLMs remain unreliable at solving problems about, or even interpreting mathematical visuals. As a result, diagram judgments made by LLMs or MLMs can be inconsistent, mathematically incorrect, or misaligned with pedagogical intent. And notably, closed-source and larger frontier models have been shown to be more effective at LLM-as-a-Judge techniques \cite{zheng2023judging}. However, their high costs and limited accessibility make them impractical for real-world educational deployments, especially in resource-constrained settings. For education technology to be truly equitable, evaluation methods must work reliably with open models and lightweight pipelines that can be deployed at scale at low costs.

This gap highlights the need for task-specific, automated pipelines for evaluating mathematical diagrams. In our work, we present \approachname: an automatic pipeline that evaluates TikZ-generated diagrams using back-translation \cite{sennrich2016improving, edunov2018understanding, fadaee2018backtranslation} and grammar-based mathematical and spatial checks. Specifically, our contributions are as follows:
\begin{enumerate}
    \item We introduce a back-translation pipeline to automatically and scalably evaluate mathematical and spatial criteria of diagrams (Section~\ref{sec:methods_ir_and_backtranslation}). This novel method outperforms LLM-as-a-Judge on agreement with human evaluation, with stronger performance across all models, which is essential for educational deployments (Section \ref{sec:results}).
    \item We utilize a 398-item evaluation dataset of real-world mathematical diagram generations from teachers (Section \ref{sec:methods_dataset}), demonstrating an approach for the continued development and improvement of automatic evaluation pipelines for mathematical diagrams. 
\end{enumerate}

\begin{figure}
    \centering
    \includegraphics[width=0.8\linewidth]{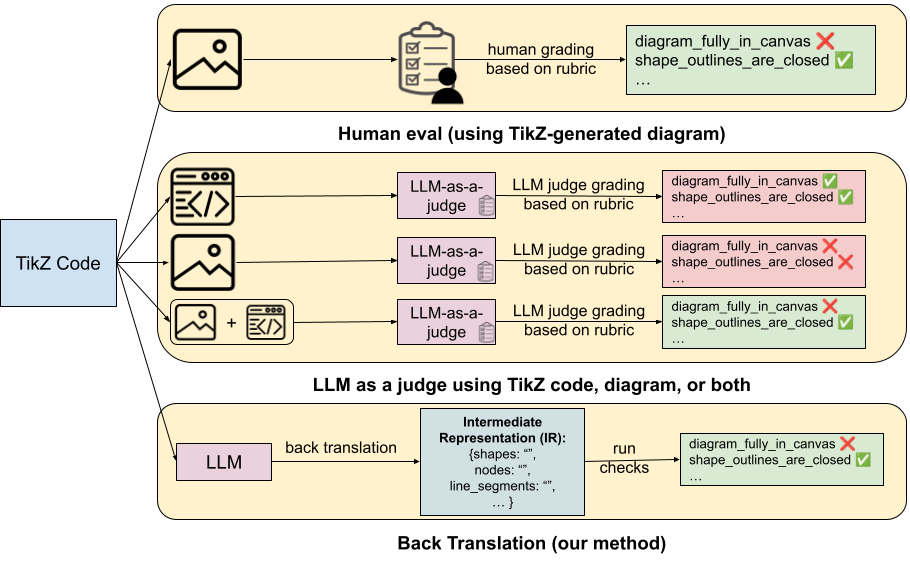}
    \caption{Different evaluation approaches for TikZ-generated code. 
    The left shows TikZ code as the common input. 
    \textbf{Top:} We asked human evaluators to rate the TikZ compiled images based on the rubric discussed in Section \ref{sec:methods_rubric}. 
    \textbf{Middle:} LLM-as-a-Judge uses either the TikZ code, the rendered image, or both to make judgments. 
    \textbf{Bottom:} In our back-translation method, an LLM translates the TikZ code into an intermediate representation (IR) with multiple fields, after which automatic checks (e.g., whether the diagram is fully in canvas or whether outlines are closed) are run. A diagram is considered valid if all check pass.}
    \label{fig:full_pipeline}
    \vspace{-1em}
\end{figure}
\vspace{-1em}
\section{Methods}
\vspace{-0.5em}
\subsection{Evaluation Dataset - Geometric Diagrams}
\label{sec:methods_dataset}
Given the downstream educational application of our pipeline, we ground our evaluation dataset in real-world data. Specifically, we use conversational data from Coteach\footnote{\url{https://coteach.ai/}}, an AI assistant for mathematics educators that uses the Illustrative Mathematics K–12 Math v.360 curriculum. For the dataset to reflect the types of diagrams teachers most commonly generate, we pulled $6,000$ random conversations between teachers and the tool. The most frequent type of diagram request category were geometric constructions, which is why we choose to focus on 2D and 3D shapes in this work.

% vishal can you add number specifics on subgroups
We sampled $398$ instances of geometric constructions (i.e. pairs of (teacher's request, generated code)), resulting in $208$ 2D diagrams (triangles, circles, and rectangles) and $190$ 3D shape diagrams (prisms and cubes). We utilize $12$ diagrams for human-eval calibration and pipeline development, leaving $386$ for our test set. Since our entire dataset is derived from natural teacher-LLM interactions, we can evaluate the quality of diagrams used ``in the wild''. The diagrams in the dataset represent a variety of error types (see Appendix~\ref{app:dataset_errs}, which informed the design of our evaluation rubric.

\subsection{Rubric Construction}
\label{sec:methods_rubric}
We developed a rubric that aims to capture criteria that (i) correspond to the most frequent failure modes observed in generated diagrams, (ii) are unambiguous, observable, and (iii) compatible with both human and LLM judging. The rubric targets two axes: \emph{mathematical correctness} and \emph{spatial correctness}. We designed the rubric by first reviewing a small sample of Coteach diagrams and open-coding recurrent failure modes (off-frame content, mislabeled angles, etc.). Then, we iteratively refined the criteria through calibration rounds where we scored parts of the Coteach dataset using our rubric to identify gaps. Our final rubric is shown in Table \ref{tab:rubric}.
\begin{table}[htbp]
\centering
\small
\begin{tabular}{p{0.52\linewidth} p{0.36\linewidth}}
\toprule
\textbf{Criterion} & \textbf{Allowed values} \\
\midrule
\textbf{Mathematical correctness} & \\
% Shape is closed (no gaps in outline) & Yes \;|\; No \\
Labeled angles match drawn angles & Yes \;|\; No \;|\; N/A \\
Labeled lengths/areas match proportions & Yes \;|\; No \;|\; N/A \\
% Core mathematical properties of the shape are correct & Yes \;|\; No \\
\textbf{Spatial correctness} & \\
Diagram is fully in frame & Yes \;|\; No \\
Elements are scaled to be readable & Yes \;|\; No \\
Labels are associated with correct elements & Yes \;|\; No \;|\; N/A \\
Elements do not problematically overlap & Yes \;|\; No \\
\bottomrule
\end{tabular}
\caption{The rubric created for assessing the mathematical and spatial correctness of math diagrams. The rubric was used for human annotation, and was mirrored in the automated pipeline.}
\label{tab:rubric}
\end{table}

\begin{table}[t]
\centering
\small

\begin{tabular}{l|ccc|ccc}
\toprule
\textbf{Models} & \multicolumn{3}{c|}{\textbf{Back-translation}} & \multicolumn{3}{c}{\textbf{LLM-as-a-Judge (code+image)}} \\
\cmidrule(lr){2-4} \cmidrule(lr){5-7}
 & Cohen's $\kappa$ &Time (s)& Cost (\$) & Cohen's $\kappa$  & Time (s)& Cost (\$)\\
\midrule
GPT-4.1        & \textbf{0.562} & 25.78 & 6.75 & 0.399 &  14.50 & 3.61 \\
GPT-5          & \textbf{0.555}  & 36.46 & 10.29 & 0.498 &  26.20 & 4.85 \\
GPT-4.1 Mini   & \textbf{0.483} & 12.16 & \textbf{0.47} & 0.388 & 8.55 & 0.82 \\
GPT-5 Mini     & \textbf{0.527}  & 42.14 & 2.12 & 0.465 &  12.66 & 0.86 \\
% LLaMA-4 Scout  &   &   &   &   &   &   &   &   \\
% LLaMA-4 Maverick &   &   &   &   &   &   &   &   \\
% Gemma 12B      &   &   &   &   &   &   &   \\
% Gemma 27B      &   &   &   &   &   &   &   \\
\bottomrule
\end{tabular}
\caption{Comparison of models across two evaluation approaches. 
For each method, \textbf{Cohen's $\kappa$} measures agreement with human ratings, 
\textbf{Time} reports the avg time required to evaluate a diagram in the dataset, 
and \textbf{Cost} reports the total API cost of evaluation for the dataset. 
Back-translation evaluates diagrams by using an LLM to translate the TikZ code to an \emph{Intermediate Representation}, which programmatic checks are run on,
while LLM-as-a-Judge evaluates diagrams from their code and image. We provide similar comparison tables using LLM-as-a-Judge provided only with TikZ code and with image in Appendix \ref{app:add_results}.}
\label{tab:model_comparison}
\vspace{-2em}
\end{table}

\vspace{-0.5em}
\subsection{Intermediate Representations and Back Translation}
\label{sec:methods_ir_and_backtranslation}
To represent the key features of mathematical diagrams in a generalizable manner, we build an intermediate representation (IR) of the diagrams. Our use of an IR is inspired by compiler design, where source code is first translated into a structured, machine-interpretable form before further analysis and optimization \cite{lattner2004llvm, alfred2007compilers}. Similarly, by mapping TikZ into a standardized IR of shapes and relationships, we decouple varied formatting differences and drawing patterns from deterministic verification, allowing rule-based checks to operate on a 1:1 mapping between diagram entities and IR fields. The IR schema and example diagrams and their IRs are available in Appendix \ref{app:ir_examples}. % david, what are some good papers to cite here on the compiler point?

The core idea behind our pipeline relies on \emph{back-translation}, an idea inspired by its counterpart in Neural-Machine Translation, which entails translation from the target language (in our case, TikZ code) to the source language (the intermediate representation). This enables the mapping of high-entropy TikZ into a low-entropy, schema-constrained ``pivot language'' (our IR) where deterministic checks are easy to write. By back-translating into an IR, we (i) factor out surface form (code formatting, macro usage), (ii) localize errors to explicit types, (iii) enable validators that run without a large model, and (iv) gain auditability and reproducibility, since the checks explain precisely why a diagram passes or fails. Conceptually, this mirrors back-translation in NMT \cite{sennrich2016improving, edunov2018understanding, fadaee2018backtranslation}: information is projected into a representation where constraints are easier to enforce. Practically, it decouples perception from verification: an LLM performs the semantic analysis (TikZ→IR), while rule-based checks perform the evaluation task~\ref{app:check_pseudocodes}.

\vspace{-0.5em}
\subsection{LLM-as-a-Judge}
\label{sec:methods_llmasjudge}
We compare our pipeline to an LLM-as-a-Judge pipeline (including both LLMs and MLMs) that asks the LLM to act as an impartial judge and evaluate each of the rubric criteria for three conditions: just the diagram, just the TikZ code, and both the diagram and the TikZ code. For all LLM-as-a-Judge conditions, we set temperature to 0 and and top-p to 1. For GPT-5, we used the “low reasoning” mode to control latency. The prompt used can be found in Appendix \ref{app:llmasjudge_prompt}.

\vspace{-0.5em}
\section{Results}
\label{sec:results}
\vspace{-0.5em}
Table \ref{tab:model_comparison} compares back-translation with LLM-as-a-Judge across four models. We find that back-translation outperforms LLM-as-a-Judge in its strongest setting (where it uses both code and image input), demonstrating   comparable agreement with human raters $(\kappa = 0.48 - 0.56)$\citep{cohen1960kappa} while LLM-as-a-Judge shows weaker agreement $(\kappa = 0.39 - 0.47)$. However, because back-translation decouples parsing the TikZ code from verifying its correctness, even the weakest model, GPT-4.1-Mini, demonstrates similar performance with back-translation as compared to the best LLM-judge (GPT-5) at 10.3x lower the cost ($\$0.47$ vs. $\$4.83$). This is notable, given that for AI-based education tools to be accessible and scalable, they need to be built at the lowest cost possible. 
 % While LLM-as-a-Judge attains slightly higher Micro-F1, back-translation consistently delivers stronger human agreement at lower cost. 

In Table~\ref{tab:checkwise_vs_llm_judge_both}, we compare the performance of back-translation for specific checks with that of LLM-as-a-Judge supplied with an image and code. We measure Cohen's $\kappa$ against human ratings. We see that for most checks, back-translation results in higher human agreement than LLM-as-a-judge. Notably, back-translation performs better for both spatial checks, despite the LLM-as-a-judge being provided the image to judge with. LLM-as-a-judge considerably outperforms backtranslation for the mathematical check about angle labels (0.829 vs. 0.652). This is likely because programmatically checking the positioning of angle labels in relation to surrounding geometric objects is tricky. However, given an image, it is easy to check whether or not an angle is labeled properly.
\begin{table}[htbp]
\centering
\small
\setlength{\tabcolsep}{4pt}
\renewcommand{\arraystretch}{0.9}
\begin{tabular}{p{3cm}cccccccc}
\toprule
 & \multicolumn{4}{c}{Backtranslation ($\kappa$)} & \multicolumn{4}{c}{Judge, image + code ($\kappa$)} \\
\cmidrule(lr){2-5} \cmidrule(lr){6-9}
Criterion & gpt-4.1 & gpt-4.1 mini & gpt-5 & gpt-5 mini & gpt-4.1 & gpt-4.1 mini & gpt-5 & gpt-5 mini \\
\midrule
\textbf{Mathematical checks} \\
Labeled angles match drawn angle & 0.691 & 0.666 & 0.644 & 0.652 & 0.793 & 0.791 & 0.795 & \textbf{0.829} \\
Lengths/areas match proportions & 0.449 & 0.480 & 0.429 & 0.422 & 0.581 & 0.596 & \textbf{0.673} & 0.616 \\
\textbf{Spatial checks} \\
Diagram fully in frame & 0.573 & 0.397 & \textbf{0.604} & 0.581 & 0.184 & 0.162 & 0.390 & 0.398 \\
Elements scaled to be readable & \textbf{0.362} & 0.283 & 0.334 & 0.272 & -0.017 & 0.000 & 0.043 & 0.097 \\
Labels associated with elements & \textbf{0.812} & 0.697 & 0.715 & 0.687 & 0.789 & 0.780 & 0.768 & 0.757 \\
No problematic overlap & 0.489 & 0.377 & \textbf{0.608} & 0.549 & 0.062 & 0.000 & 0.315 & 0.094 \\
\bottomrule
\end{tabular}
\caption{Rubric check-wise comparison between back-translation and LLM-as-a-judge supplied with both image and code. For other LLM-as-a-Judge settings, we include results in Appendix~\ref{app:rubric_checkwise}}.
\label{tab:checkwise_vs_llm_judge_both}
\end{table}

\vspace{-2.5em}
\section{Limitations and Conclusion}
\vspace{-1em}
\label{sec:limitations_conclusion}
In this paper we introduced \approachname, a novel approach leveraging back-translation for evaluating TikZ diagrams via an intermediate representation and rule-based checks. By decoupling the task of perception and verification (with verification being conducted by our rule-based checks), we strengthen even the smallest LLMs to better assist with diagram evaluation. Across multiple models and evaluation settings, we found that back-translation consistently achieved higher agreement with human ratings compared to LLM-as-a-Judge. Notably, it enabled smaller, low-cost models to perform on par with, or even better than, frontier models. This highlights the promise of domain-specific automatic evaluation pipelines that combine symbolic abstraction with lightweight inference, applied to critical domains, such as education.

There are several limitations to our work. First, our rubric focuses on mathematical and spatial correctness, leaving out pedagogical usefulness, which remains a critical but more subjective dimension of diagram quality. Second, the intermediate representation captures a restricted set of geometric primitives and relations; more complex diagrams (e.g., multi-step constructions, coordinate plots) may require iterating upon the schema and checks. Third, back-translation relies on LLMs for parsing TikZ into IRs, which introduced stochastic errors during the IR generation step. While our results suggest that even smaller models perform competitively, fine-tuning a small model specifically for TikZ$\rightarrow$IR translation could further reduce costs and improve reliability. Finally, our dataset is grounded in one curriculum (Illustrative Mathematics), and additional validation on other domains (e.g., physics diagrams, freehand sketches) is needed to establish broader generalizability.

Future work should explore extending the rubric toward pedagogical dimensions, expanding the IR to cover a broader set of diagram constructs, and integrating the method into diagram-generation tools. \nope{Our results suggest that leveraging structured intermediate representations is a promising direction for building scalable, equitable, and interpretable evaluation pipelines in education technology.}

\section{Acknowledgments}
This work would not have been possible without the support of the Coteach team from Teaching Lab Studio. We are especially grateful to Peter Edmonds for providing access to the Coteach data that helped shape our dataset. We also thank the Gates Foundation (Grant \#068816) and the Stanford Institute for Human-Centered AI for funding this work. For SM and DB, this work was partially supported by the Wallenberg AI, Autonomous Systems and Software Program (WASP) funded by the Knut and Alice Wallenberg Foundation.

\bibliography{references}

\clearpage

\setcounter{section}{0}
\appendix  
\renewcommand{\thesection}{A\arabic{section}}

% Create & print a TOC that includes ONLY entries added after this point
\startcontents[appendix]
\printcontents[appendix]{l}{1}{\section*{Appendix}}

\section{Additional Results}
\label{app:add_results}
\subsection{Cohen's $\kappa$, time, and cost for other LLM-as-a-Judge settings}
\begin{table}[!htbp]
\centering
\begin{tabular}{l|ccc|ccc}
\toprule
\textbf{Models} & \multicolumn{3}{c|}{\textbf{Back-translation}} & \multicolumn{3}{c}{\textbf{LLM-as-a-Judge (code)}} \\
\cmidrule(lr){2-4} \cmidrule(lr){5-7}
 & Cohen's $\kappa$ & Time (s) & Cost (\$) & Cohen's $\kappa$ & Time (s) & Cost (\$) \\
\midrule
GPT-4.1        & \textbf{0.563} & 25.74 & 6.76 & 0.395 & 9.13  & 3.30 \\
GPT-5          & \textbf{0.556} & 36.93 & 10.30 & 0.427 & 25.32 & 5.85 \\
GPT-4.1 Mini   & \textbf{0.483} & 12.69 & 0.47 & 0.388 & 10.25 & 0.69 \\
GPT-5 Mini     & \textbf{0.527} & 42.26 & 2.25 & 0.406 & 12.77 & 0.84 \\
\bottomrule
\end{tabular}
\caption{Back-translation vs LLM-as-a-Judge (code).}
\label{tab:model_comparison_code}
\end{table}

\begin{table}[!htbp]
\centering
\begin{tabular}{l|ccc|ccc}
\toprule
\textbf{Models} & \multicolumn{3}{c|}{\textbf{Back-translation}} & \multicolumn{3}{c}{\textbf{LLM-as-a-Judge (image)}} \\
\cmidrule(lr){2-4} \cmidrule(lr){5-7}
 & Cohen's $\kappa$ & Time (s) & Cost (\$) & Cohen's $\kappa$ & Time (s) & Cost (\$) \\
\midrule
GPT-4.1        & \textbf{0.563} & 25.74 & 6.76 & 0.365 & 12.28 & 2.77 \\
GPT-5          & \textbf{0.556} & 36.93 & 10.30 & 0.442 & 19.37 & 3.75 \\
GPT-4.1 Mini   & \textbf{0.483} & 12.69 & 0.47 & 0.366 & 7.29  & 0.64 \\
GPT-5 Mini     & \textbf{0.527} & 42.26 & 2.25 & 0.442 & 10.88 & 0.72 \\
\bottomrule
\end{tabular}
\caption{Comparison of models: Back-translation vs LLM-as-a-Judge (image).}
\label{tab:model_comparison_image}
\end{table}

\subsection{Rubric check-wise comparison for other LLM-as-a-judge settings}
\label{app:rubric_checkwise}
In Tables~\ref{tab:checkwise_vs_llm_judge_code} and ~\ref{tab:checkwise_vs_llm_judge_image} we include results for checkwise comparisons of back-translation vs. LLM-as-a-judge (code) and LLM-as-a-judge (image), respectively.

\begin{table}[htbp]
\centering
\small
\setlength{\tabcolsep}{2pt}
\renewcommand{\arraystretch}{0.95}
\begin{tabular}{lcccccccc}
\toprule
\multirow{2}{*}{Rubric check} &
\multicolumn{4}{c}{Backtranslation} &
\multicolumn{4}{c}{Judge – code} \\
\cmidrule(lr){2-5} \cmidrule(lr){6-9}
 & GPT-4.1 & GPT-4.1 Mini & GPT-5 & GPT-5 Mini &
   GPT-4.1 & GPT-4.1 Mini & GPT-5 & GPT-5 Mini \\
\midrule
Angles labels    & 0.691 & 0.666 & 0.644 & 0.652 & \textbf{0.847} & 0.778 & 0.797 & 0.846 \\
Lengths/areas & 0.449 & 0.480 & 0.429 & 0.422 & 0.581 & 0.615 & \textbf{0.660} & 0.619 \\
Diagram in frame               & 0.573 & 0.397 & \textbf{0.604} & 0.581 & 0.161 & 0.172 & 0.230 & 0.197 \\
Readable size       & \textbf{0.362} & 0.283 & 0.334 & 0.272 & 0.000 & -0.005 & -0.005 & -0.005 \\
Labels associated     & \textbf{0.812} & 0.697 & 0.715 & 0.687 & 0.781 & 0.771 & 0.773 & 0.782 \\
Problematic overlap & 0.489 & 0.377 & \textbf{0.608} & 0.549 & 0.000 & 0.000 & 0.105 & -0.005 \\
\bottomrule
\end{tabular}
\caption{$\kappa$ comparison: Backtranslation vs. LLM-as-a-Judge (code).}
\label{tab:checkwise_vs_llm_judge_code}
\end{table}
\FloatBarrier

\begin{table}[!htbp]
\centering
\small
\setlength{\tabcolsep}{2pt}
\renewcommand{\arraystretch}{0.95}
\begin{tabular}{lcccccccc}
\toprule
\multirow{2}{*}{Rubric check} &
\multicolumn{4}{c}{Backtranslation} &
\multicolumn{4}{c}{Judge – image} \\
\cmidrule(lr){2-5} \cmidrule(lr){6-9}
 & GPT-4.1 & GPT-4.1 Mini & GPT-5 & GPT-5 Mini &
   GPT-4.1 & GPT-4.1 Mini & GPT-5 & GPT-5 Mini \\
\midrule
Angles labels     & 0.691 & 0.666 & 0.644 & 0.652 & 0.706 & 0.703 & \textbf{0.738} & 0.718 \\
Lengths/areas  & 0.449 & 0.480 & 0.429 & 0.422 & \textbf{0.597} & 0.593 & 0.583 & 0.595 \\
Diagram in frame               & 0.573 & 0.397 & \textbf{0.604} & 0.581 & 0.102 & 0.140 & 0.313 & 0.242 \\
Readable size       & \textbf{0.362} & 0.283 & 0.334 & 0.272 & -0.024 & -0.005 & 0.218 & 0.334 \\
Labels associated     & \textbf{0.812} & 0.697 & 0.715 & 0.687 & 0.776 & 0.765 & 0.661 & 0.727 \\
Problematic overlap & 0.489 & 0.377 & \textbf{0.608} & 0.549 & 0.034 & 0.000 & 0.140 & 0.034 \\
\bottomrule
\end{tabular}
\caption{$\kappa$ comparison: Backtranslation vs. LLM-as-a-Judge (image).}
\label{tab:checkwise_vs_llm_judge_image}
\end{table}
\FloatBarrier

\subsection{Confusion matrices}
In Tables~\ref{tab:gpt-5_confmat} through ~\ref{tab:gpt-41-mini_confmat}, for each rubric check, we include the number of true positives, true negatives, false positives (overly-cautious rule-based checks that mark something as incorrect when humans marked it as correct), and false negatives (rule-based checks that mark a pass where humans found a failure).

\begin{table}[!htbp]
\centering
\small
\setlength{\tabcolsep}{6pt}
\begin{tabular}{lcccc}
\toprule
\textbf{Criterion} & TP & TN & FP & FN \\
\midrule
\textbf{Mathematical correctness} & \\
Labeled lengths/areas match proportions & 33 (8.5\%) & 53 (13.7\%) & 38 (9.8\%) & 17 (4.4\%) \\
Labeled angles match drawn angles & 4 (1.0\%) & 28 (7.3\%) & 8 (2.1\%) & 6 (1.6\%) \\
\textbf{Spatial correctness} & \\
Diagram fully in frame & 32 (8.3\%) & 320 (82.9\%) & 25 (6.5\%) & 9 (2.3\%) \\
Elements are scaled to be readable & 5 (1.3\%) & 363 (94.0\%) & 6 (1.6\%) & 12 (3.1\%) \\
Labels associated with correct elements & 11 (2.8\%) & 188 (48.7\%) & 3 (0.8\%) & 29 (7.5\%) \\
Elements do not problematically overlap & 61 (15.8\%) & 272 (70.5\%) & 25 (6.5\%) & 28 (7.3\%) \\
\midrule
\textbf{Total (all applicable)} & 152 (6.5\%) & 1888 (80.4\%) & 185 (7.9\%) & 123 (5.2\%) \\
\bottomrule
\end{tabular}
\caption{Per-criterion confusion breakdown for GPT-5 against human evaluation. TP=both mark the check as failing, TN=both mark it as passing, FP=model marks a failure humans do not, FN=model marks a pass where humans found a failure. Row percentages use $N=386$ diagrams; totals are normalized by applicable slots ($N=2348$), since some criteria do not apply to every diagram (e.g., angle labels).}
\label{tab:gpt-5_confmat}
\end{table}
\FloatBarrier

\begin{table}[!htbp]
\centering
\small
\setlength{\tabcolsep}{6pt}
\begin{tabular}{lcccc}
\toprule
\textbf{Criterion} & TP & TN & FP & FN \\
\midrule
\textbf{Mathematical correctness} & \\
Labeled lengths/areas match proportions & 34 (8.8\%) & 52 (13.5\%) & 39 (10.1\%) & 16 (4.1\%) \\
Labeled angles match drawn angles & 5 (1.3\%) & 28 (7.3\%) & 8 (2.1\%) & 5 (1.3\%) \\
\textbf{Spatial correctness} & \\
Diagram fully in frame & 33 (8.5\%) & 314 (81.3\%) & 31 (8.0\%) & 8 (2.1\%) \\
Elements are scaled to be readable & 6 (1.6\%) & 361 (93.5\%) & 8 (2.1\%) & 11 (2.8\%) \\
Labels associated with correct elements & 12 (3.1\%) & 182 (47.2\%) & 9 (2.3\%) & 28 (7.3\%) \\
Elements do not problematically overlap & 58 (15.0\%) & 255 (66.1\%) & 42 (10.9\%) & 31 (8.0\%) \\
\midrule
\textbf{Total (all applicable)} & 151 (6.4\%) & 1904 (81.1\%) & 169 (7.2\%) & 124 (5.3\%) \\
\bottomrule
\end{tabular}
\caption{Per-criterion confusion breakdown for GPT-4.1 against human evaluation. TP=both mark the check as failing, TN=both mark it as passing, FP=model marks a failure humans do not, FN=model marks a pass where humans found a failure. Row percentages use $N=386$ diagrams; totals are normalized by applicable slots ($N=2348$), since some criteria do not apply to every diagram (e.g., angle labels).}
\label{tab:gpt-41_confmat}
\end{table}
\FloatBarrier

\begin{table}[!htbp]
\centering
\small
\setlength{\tabcolsep}{6pt}
\begin{tabular}{lcccc}
\toprule
\textbf{Criterion} & TP & TN & FP & FN \\
\midrule
\textbf{Mathematical correctness} & \\
Labeled lengths/areas match proportions & 32 (8.3\%) & 53 (13.7\%) & 38 (9.8\%) & 18 (4.7\%) \\
Labeled angles match drawn angles & 5 (1.3\%) & 26 (6.7\%) & 10 (2.6\%) & 5 (1.3\%) \\
\textbf{Spatial correctness} & \\
Diagram fully in frame & 31 (8.0\%) & 319 (82.6\%) & 26 (6.7\%) & 10 (2.6\%) \\
Elements are scaled to be readable & 5 (1.3\%) & 358 (92.7\%) & 11 (2.8\%) & 12 (3.1\%) \\
Labels associated with correct elements & 13 (3.4\%) & 185 (47.9\%) & 6 (1.6\%) & 27 (7.0\%) \\
Elements do not problematically overlap & 60 (15.5\%) & 263 (68.1\%) & 34 (8.8\%) & 29 (7.5\%) \\
\midrule
\textbf{Total (all applicable)} & 150 (6.4\%) & 1872 (79.7\%) & 201 (8.6\%) & 125 (5.3\%) \\
\bottomrule
\end{tabular}
\caption{Per-criterion confusion breakdown for GPT-5-mini against human evaluation. TP=both mark the check as failing, TN=both mark it as passing, FP=model marks a failure humans do not, FN=model marks a pass where humans found a failure. Row percentages use $N=386$ diagrams; totals are normalized by applicable slots ($N=2348$), since some criteria do not apply to every diagram (e.g., angle labels).}
\label{tab:gpt-5-mini_confmat}
\end{table}
\FloatBarrier

\begin{table}[!htbp]
\centering
\small
\setlength{\tabcolsep}{6pt}
\begin{tabular}{lcccc}
\toprule
\textbf{Criterion} & TP & TN & FP & FN \\
\midrule
\textbf{Mathematical correctness} & \\
Labeled lengths/areas match proportions & 32 (8.3\%) & 44 (11.4\%) & 47 (12.2\%) & 18 (4.7\%) \\
Labeled angles match drawn angles & 5 (1.3\%) & 25 (6.5\%) & 11 (2.8\%) & 5 (1.3\%) \\
\textbf{Spatial correctness} & \\
Diagram fully in frame & 26 (6.7\%) & 303 (78.5\%) & 42 (10.9\%) & 15 (3.9\%) \\
Elements are scaled to be readable & 5 (1.3\%) & 359 (93.0\%) & 10 (2.6\%) & 12 (3.1\%) \\
Labels associated with correct elements & 10 (2.6\%) & 178 (46.1\%) & 13 (3.4\%) & 30 (7.8\%) \\
Elements do not problematically overlap & 44 (11.4\%) & 259 (67.1\%) & 38 (9.8\%) & 45 (11.7\%) \\
\midrule
\textbf{Total (all applicable)} & 125 (5.3\%) & 1872 (79.7\%) & 201 (8.6\%) & 150 (6.4\%) \\
\bottomrule
\end{tabular}
\caption{Per-criterion confusion breakdown for GPT-4.1-mini against human evaluation. TP=both mark the check as failing, TN=both mark it as passing, FP=model marks a failure humans do not, FN=model marks a pass where humans found a failure. Row percentages use $N=386$ diagrams; totals are normalized by applicable slots ($N=2348$), since some criteria do not apply to every diagram (e.g., angle labels).}
\label{tab:gpt-41-mini_confmat}
\end{table}
\FloatBarrier

\section{Additional Details on Methods}

\subsection{Dataset Diagram Error Distributions as Determined by Human Evaluation}
In Table~\ref{tab:rubric_distribution}, we show the frequency of errors represented in our dataset as measured by human evaluation. 
\label{app:dataset_errs}
\begin{table}[htbp]
\centering
\small
\setlength{\tabcolsep}{6pt}
\renewcommand{\arraystretch}{1.1}
\begin{tabular}{lrrr}
\toprule
Rubric check & Yes & No & N/A \\
\midrule
\multicolumn{4}{l}{\textbf{\textit{Mathematical checks}}} \\
Labeled lengths/areas match proportions & 91 (23.6\%) & 50 (13.0\%) & 245 (63.5\%) \\
Angle labels match arcs                 & 36 (9.3\%)  & 10 (2.6\%)  & 340 (88.1\%) \\
\midrule
\multicolumn{4}{l}{\textbf{\textit{Spatial checks}}} \\
Diagram fully in canvas                 & 345 (89.4\%) & 41 (10.6\%) & -- \\
Elements do not problematically overlap & 297 (77.0\%) & 89 (23.1\%) & -- \\
Elements are readable size              & 369 (95.6\%) & 17 (4.4\%)  & -- \\
Labels associated with correct elements         & 191 (49.5\%) & 40 (10.4\%) & 155 (40.2\%) \\
\bottomrule
\end{tabular}
\caption{Distribution of rubric outcomes (counts with percentages), grouped by mathematical and spatial checks.}
\label{tab:rubric_distribution}
\end{table}
\FloatBarrier

\subsection{Intermediate Representation (IR) Examples}
\label{app:ir_examples}
\begin{figure}[!htbp]
    \centering
    \includegraphics[width=\linewidth]{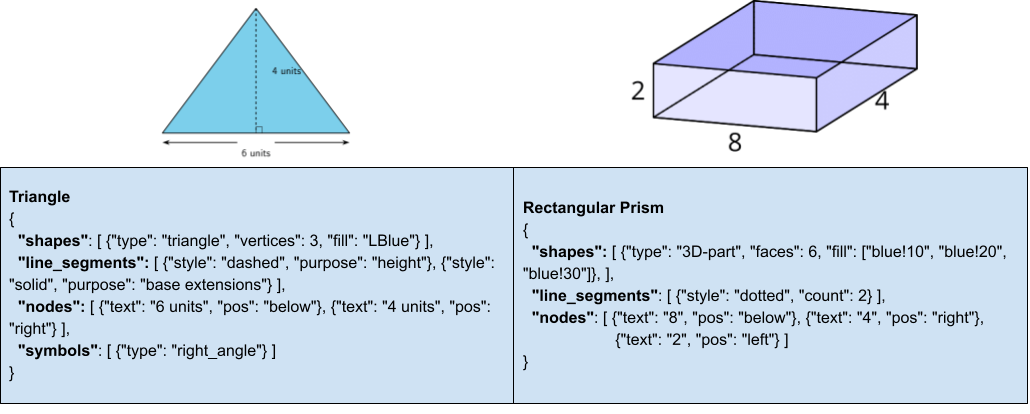}
    \caption{Examples of abbreviated intermediate representations (IRs) extracted 
    from TikZ diagrams. The triangle (left) and rectangular prism (right) illustrate 
    how diagrams are mapped into structured IRs of shapes, line segments, nodes, 
    and symbols. For clarity, only key fields are shown here; the full IR schema 
    and detailed descriptions of all attributes are provided in the section below.}
\end{figure}
\FloatBarrier

\subsection{IR Schema and Attributes}
\label{app:ir_schema}

The IR schema defines a structure for the key properties of shape diagrams. 

\begin{lstlisting}[
    language=Python,
    caption={Intermediate Representation},
    label={lst:tikz-ir-schema},
    basicstyle=\small\ttfamily,
    breaklines=true,
    showstringspaces=false,
    numbers=none,
    stepnumber=0
]
# Core type definitions
Coord2D = conlist(float, min_length=2, max_length=2)  # [x, y]
Coord3D = conlist(float, min_length=3, max_length=3)  # [x, y, z]
Coord2Dor3D = Union[Coord2D, Coord3D]

# Core classes, others omitted for brevity
class Transform(BaseModel):
    shift: Optional[Coord2Dor3D] = None
    scale: Optional[Union[float, List[float]]] = None
    rotate: Optional[float] = None
    # xshift, yshift omitted for brevity

class Node(BaseModel):
    at: Coord2Dor3D
    text: str
    anchor: Optional[Literal['mid', 'above', 'below', 'left', 'right',
                            'above left', 'above right', 'below left', 
                            'below right']] = None
    transform: Optional[Transform] = None
    fill: Optional[str] = None

class Shape2D(BaseModel):
    vertices: List[Coord2D]
    type: Literal['triangle', 'polygon']
    cycle: bool
    transform: Optional[Transform] = None
    fill: Optional[str] = None

class Shape3DPart(BaseModel):
    type: Literal['3D-part']
    id: int
    vertices: List[Coord3D]
    cycle: bool
    transform: Optional[Transform] = None
    fill: Optional[str] = None

class LineSegment(BaseModel):
    from_: Coord2Dor3D = Field(alias='from')
    to: Coord2Dor3D
    style: Optional[Literal['solid', 'dashed', 'dotted']] = 'solid'
    text: Optional[str] = None
    transform: Optional[Transform] = None

class Arc(BaseModel):
    center: Coord2Dor3D
    start_angle: float
    end_angle: float
    radius: float
    transform: Optional[Transform] = None
    fill: Optional[str] = None

# Main IR structure
class TikzIR(BaseModel):
    tikzpicture_options: Optional[CoordinateSystem] = None
    clips: Optional[List[Clip]] = None
    shapes: Optional[List[Union[Shape2D, Shape3DPart, Ucube]]] = None
    line_segments: Optional[List[LineSegment]] = None
    nodes: Optional[List[Node]] = None
    circles: Optional[List[Circle]] = None
    rectangle_primitives: Optional[List[RectanglePrimitive]] = None
    arcs: Optional[List[Arc]] = None
\end{lstlisting}

The schema consists of a main \texttt{TikzIR} class containing optional lists of geometric primitives (nodes, shapes, line segments, arcs, etc.), each with their own coordinate specifications and optional transformation parameters. 

\subsection{Rule-based check pseudo code}
\label{app:check_pseudocodes}

% 1. Labeled lengths/areas match proportions
\begin{algorithm}[!htbp]
\caption{Check: Labeled lengths/areas match proportions}
\begin{algorithmic}[1]
\State Extract all numeric labels from IR nodes and segment annotations
\State Collect all line segments and polygons (shapes + rectangles)
\For{each numeric label $L$ with value $v$}
    \State Find nearest segment $S$ and distance $d_S$
    \State Find containing polygon $P$ 
    \If{$d_S \leq 12$pt}
        \State Associate $L$ with segment $S$ as a length label
    \ElsIf{$L$ inside $P$}
        \State Associate $L$ with polygon $P$ as an area label
    \ElsIf{$d_S < \infty$}
        \State Associate $L$ with segment $S$ as a length label
    \EndIf
\EndFor
\For{each pair of length labels $(L_i, L_j)$ on segments $(S_i, S_j)$}
    \State $m_i \gets$ measured length of $S_i$, $m_j \gets$ measured length of $S_j$
    \If{$|m_i \cdot v_j - m_j \cdot v_i| > \epsilon \cdot \max(m_i \cdot v_j, m_j \cdot v_i)$}
        \State Report mismatch
    \EndIf
\EndFor
\State (Repeat pairwise check for area labels on polygons)
\end{algorithmic}
\end{algorithm}
\FloatBarrier

% 2. Labeled angles match drawn angles
\begin{algorithm}[!htbp]
\caption{Check: Labeled angles match drawn angles}
\begin{algorithmic}[1]
\State Extract all arcs and their geometric properties from IR
\State Extract all angle labels (nodes containing degree symbols)
\If{no arcs and no angle labels}
    \State \Return N/A
\EndIf
\For{each angle label $L$ with text containing numeric value $\theta_L$}
    \State Find nearest arc $A$ by geometric distance
    \State $d \gets$ distance from $L$ to $A$
    \State $\tau \gets$ adaptive tolerance based on arc size
    \If{$d > \tau$}
        \State Report label too far from arc
        \State \textbf{continue}
    \EndIf
    \State Parse numeric angle $\theta_L$ from label text
    \State Compute arc sweep angle $\theta_A$ 
    \If{$|\theta_A - \theta_L| > \epsilon$}
        \State Report angle mismatch
    \EndIf
\EndFor
\State Check right-angle symbols (small squares) against 90° angles
\end{algorithmic}
\end{algorithm}
\FloatBarrier

% 3. Diagram fully in frame
\begin{algorithm}[!htbp]
\caption{Check: Diagram fully in frame}
\begin{algorithmic}[1]
\State Initialize working canvas from page bounds or clip regions in IR
\If{working canvas is empty}
    \State \Return FAIL
\EndIf
\State Buffer canvas by tolerance $\tau$ (default: 2pt)
\For{each geometric entity $E$ in IR (shapes, circles, segments, arcs)}
    \State Convert $E$ to geometry $G$ using coordinate system
    \If{$G \not\subseteq$ working canvas}
        \State Report $E$ exceeds canvas bounds
        \State \Return FAIL
    \EndIf
\EndFor
\For{each text node $N$ with non-empty text}
    \State Convert $N$ to bounding box geometry $G$
    \If{$G \not\subseteq$ working canvas}
        \State Report node exceeds canvas bounds
        \State \Return FAIL
    \EndIf
\EndFor
\State \Return PASS
\end{algorithmic}
\end{algorithm}
\FloatBarrier

% 4. Elements are scaled to be readable
\begin{algorithm}[!htbp]
\caption{Check: Elements are scaled to be readable}
\begin{algorithmic}[1]
\State Collect all geometric entities (shapes, circles, segments, arcs, nodes)
\State Compute overall diagram bounding box from union of all geometries
\State $d_{\text{min}} \gets \min(\text{width}, \text{height})$ of diagram bbox
\State $\tau \gets d_{\text{min}} \times r$ where $r$ is relative threshold (default: 0.02)
\For{each entity $E$ with geometry $G$}
    \If{$G$ is line-like}
        \State $m \gets$ length of $G$
    \Else
        \State $m \gets \min(\text{width}, \text{height})$ of $G$ bbox
    \EndIf
    \If{$m < \tau$}
        \State Record $E$ as undersized
    \EndIf
\EndFor
\If{any undersized elements}
    \State \Return FAIL with list of smallest elements
\Else
    \State \Return PASS
\EndIf
\end{algorithmic}
\end{algorithm}
\FloatBarrier

% 5. Labels associated with correct elements
\begin{algorithm}[!htbp]
\caption{Check: Labels associated with correct elements}
\begin{algorithmic}[1]
\State Collect shapes, segments, and arcs as candidate elements
\State Extract shape boundaries (edges) for polygon shapes
\State Collect all text nodes as labels, classify each as: angle, numeric, or text
\If{no labels present}
    \State \Return N/A
\EndIf
\For{each label $L$ with type $t$}
    \State Identify candidate elements based on type:
    \State \quad angle labels $\rightarrow$ arcs
    \State \quad numeric labels $\rightarrow$ segments and shape edges
    \State \quad text labels $\rightarrow$ all elements, prefer containment
    \State Rank candidates by (distance, priority), where priority depends on label type
    \State $C \gets$ closest candidate element
    \State $d \gets$ distance from $L$ to $C$
    \State $\tau \gets$ adaptive tolerance based on element size
    \If{$d > \tau$}
        \State Report label not associated with any element
        \State \textbf{continue}
    \EndIf
    \State Check for ties (multiple candidates within small margin)
    \State Associate $L$ with $C$
\EndFor
\State \Return PASS if all labels associated, else FAIL
\end{algorithmic}
\end{algorithm}
\FloatBarrier

% 6. Elements do not problematically overlap
\begin{algorithm}[!htbp]
\caption{Check: Elements do not problematically overlap}
\begin{algorithmic}[1]
\State Collect all text nodes with bounding boxes
\State Collect all shape boundaries, segment boundaries, arc boundaries
\State Identify 3D faces
\State \textit{(Text-text overlaps)}
\For{each pair of text nodes $(N_i, N_j)$ with different text}
    \If{bboxes intersect with area $> 0.05 \times \min(\text{area}_i, \text{area}_j)$}
        \State Report problematic text overlap
    \EndIf
\EndFor
\State \textit{(Text-geometry overlaps)}
\For{each text node $N$}
    \For{each boundary $B$ (shape edge, segment, or arc)}
        \If{$B$ intersects $N$ bbox with length $> 0.4 \times$ perimeter of $N$}
            \State Report text obscured by line
        \EndIf
    \EndFor
\EndFor
\State \textit{(3D depth ordering)}
\For{each pair of 3D faces $(F_i, F_j)$}
    \If{faces project to overlapping 2D regions}
        \State Compute mean $z$-coordinates and surface normals
        \If{depth ordering inconsistent (back face occludes front face)}
            \State Report problematic 3D overlap
        \EndIf
    \EndIf
\EndFor
\State \Return PASS if no issues, else FAIL
\end{algorithmic}
\end{algorithm}
\FloatBarrier

\subsection{Backtranslation Prompt}
\label{app:backtranslation_prompt}

\begin{lstlisting}[
    caption={TikZ to IR Backtranslation Prompt},
    label={lst:backtranslation-prompt},
    basicstyle=\footnotesize\ttfamily,
    breaklines=true,
    numbers=none,
    frame=single,
    showstringspaces=false
]
You are a deterministic parser that extracts geometric entities from TikZ code into JSON that matches the provided schema.

### Rules
1. Extract ONLY entities explicitly present in the TikZ code.
2. Omit empty fields (do not include keys with empty lists).
3. Preserve exact numerical coordinates from the code. Resolve and compute if necessary (e.g. \def or \newcommand), but DO NOT infer extra ones that are not present in the code or complete partial shapes.
4. For shapes array, list vertices in the order they are drawn, set "cycle": true if the draw command ends with `-- cycle`.
5. For rectangle_primitives array, set "is_right_angle_symbol": true when the rectangle is drawn as a right-angle marker (e.g., tiny square sharing corners with two incident edges or comments mentioning a right angle). Otherwise set it to false. If a \draw explicitly passes the 'right angle symbol' option, do not add it as a rectangle_primitive.
6. For 3D parts (shapes w/ 3D coords) and Ucubes, use one integer id per physical solid (e.g., 1, 1, 1). Faces or unit cube entries that belong to the same block---especially those emitted inside the same scope/loop---must reuse that id; only assign a new id when you are describing a genuinely different solid.
7. When you encounter the helper macro `\Ucube` (or any change of coordinates that draws the front/right/top faces of a unit cube), output a single shape object with `"type": "Ucube"` instead of three separate 3D-part entries. The cube should include its `id`, `size` (usually `[1,1,1]` unless the macro scales it), the scope transform (`scale`, `shift`, `xshift`, `yshift`), and `fill`. Do not emit the individual faces separately.
8. If a scope applies transformations (shift, scale, xshift, yshift, rotate), include them in the optional transform object. Do not numerically apply the transform.
9. Transform separation: 
   - `transform.shift` <- only TikZ's `shift={...}` argument.  
   - `transform.xshift` <- only TikZ's `xshift=...`.  
   - `transform.yshift` <- only TikZ's `yshift=...`.  
   These must NEVER be combined. If xshift/yshift values are omitted or folded into `shift`, the JSON is invalid.
10. For node options such as `rotate=...`, set the node's `node_rotate` field. Keep scope-level rotations in `transform` and do not duplicate them in `node_rotate`.
11. For tikzpicture_options, map x, y, and z to the corresponding options, or fill out 'scale' if the options include scale.
12. Loop expansion: Expand every `\foreach` loop literally. Substitute each variable with its values and emit the corresponding repeated scopes and draw commands. Never summarize or replace with a generic cube --- output must reflect exactly the iterations.
13. For custom commands, resolve the command using this table:
IM_MACROS = [
        "\TFP": "4.875in",
        "\TTP": "4.2in",
        "\TwoThirdsPage": "4.2in",
        "\HP": "3.25in",
        "\HalfPage": "3.25in",
        "\THP": "2.1in",
        "\ThirdPage": "2.1in",
        "\QP": "1.625in",
    "\QuarterPage": "1.625in",
    ]
13. Resolve relative coordinate syntax precisely: when you encounter forms like `+(...)`, `++(...)`, or `($(P)!t!(Q)$)`, evaluate them to absolute coordinates using previously defined points. For `++`, remember it updates the current point before the next coordinate is processed.

### Example

TikZ:
```latex
\foreach \x in {0,1} {
  \begin{scope}[scale=0.5, shift={(\x,0,0)}, xshift=-2in, yshift=-3in]
    \draw (0,0,0) -- (1,0,0) -- (1,1,0) -- (0,1,0) -- cycle;
  \end{scope}
}
```

JSON:

```json
{
  "shapes": [
    {
      "type": "3D-part",
      "vertices": [[0,0,0],[1,0,0],[1,1,0],[0,1,0]],
      "cycle": true,
      "id": 1,
      "transform": {
        "scale": 0.5,
        "shift": [0,0,0],
        "xshift": "-2in",
        "yshift": "-3in"
      }
    },
    {
      "type": "3D-part",
      "vertices": [[0,0,0],[1,0,0],[1,1,0],[0,1,0]],
      "cycle": true,
      "id": 2,
      "transform": {
        "scale": 0.5,
        "shift": [1,0,0],
        "xshift": "-2in",
        "yshift": "-3in"
      }
    }
  ]
}
```

### TikZ
{tikz_code}

### Output
JSON only, no explanations.
\end{lstlisting}

\subsection{LLM-as-a-Judge Prompt}
\label{app:llmasjudge_prompt}
The prompt is designed to clearly communicate the criteria and the need to thoroughly and impartially judge the diagram to allow the LLM to reason and evaluate, only differs between the three conditions in ``You are given an image of a math diagram and the LaTeX code for it'' (the other conditions including just an image or just the LaTeX code), and returns the evaluations in an analyzable format.

\begin{lstlisting}[
    frame=single,
    label={lst:llmjudge-prompt},
]
You are to act as an impartial large language model "judge". Your task is to evaluate math diagrams using the rubric provided below. You are given an image of a math diagram and the LaTeX code for it, which uses the TikZ LaTeX library. Carefully reason through the diagram's adherence to each rubric criterion before reaching any conclusions. For each diagram you review:
- Analyze and internalize the full provided diagram and rubric. 
- Systematically assess each rubric item, explaining your reasoning and specific evidence from the diagram for each, and then output your evaluation. Strictly output from the options for that rubric criterion that are provided below. 

Rubric: 
Mathematical correctness: 
Shape is closed (no gaps in outline): Yes | No - whether the diagram's shape is closed. This is independent from whether it is fully in frame (below) - is the diagram formed that it would likely be closed regardless of its framing? 
Labeled angles (if any) match the drawn angle: Yes | No | N/A - whether the labeled angles in the diagram match their labeled value. Right angle markers without a number also count. N/A if there are no labeled angles.
Labeled lengths (if any) match visual proportions: Yes | No | N/A - whether the labeled lengths or areas shown in the diagram are reasonable lengths or areas in relationship to each other. N/A if there are no labeled lengths or areas. 
Core mathematical properties of the shape are correct: Yes | No - whether the core mathematical properties of the shape are correct, independent of the criteria above. 
Spatial correctness: 
Diagram is fully in frame: Yes | No - whether all diagram elements are in the frame, and nothing is cut off.
Diagram elements are scaled to be readable: Yes | No - whether elements such as shapes, labels, etc. are sized to be readable, especially in relationship to each other.
Labels (if any) are associated with correct elements: Yes | No | N/A - whether the labels are associated with the correct elements (e.g. sides, line segments, angles, etc) in the diagram. N/A if there are no labels.
Diagram elements don't problematically overlap: Yes | No - whether no elements problematically overlap. Problematically overlapping could include labels overlapping with something so they cannot be read easily, shapes or elements of the diagram overlapping in a way that makes it challenging to interpret. A label directly intersected by a line segment would be considered problematically overlapping. 

Output Format: After reasoning and determining each criteria's evaluation, output a JSON object in the following format: 
{
  "shape_outlines_are_closed": {
    "rationale": "[Placeholder: rationale for Shape is closed (no gaps in outline)]",
    "value": "[Placeholder: Yes or No]"
  },
  "angle_labels_matches_arcs": {
    "rationale": "[Placeholder: rationale for Labeled angles (if any) match the drawn angle]",
    "value": "[Placeholder: Yes, No, or N/A]"
  },
  "labeled_lengths_areas_match_proportions": {
    "rationale": "[Placeholder: rationale for Labeled lengths (if any) match visual proportions]",
    "value": "[Placeholder: Yes, No, or N/A]"
  }
  "core_mathematical_properties_of_shapes_correct": {
    "rationale": "[Placeholder: rationale for Core mathematical properties of the shape are correct]",
    "value": "[Placeholder: Yes or No]"
  },
  "diagram_fully_in_canvas": {
    "rationale": "[Placeholder: rationale for Diagram is fully in frame]",
    "value": "[Placeholder: Yes or No]"
  },
  "diagram_elements_are_readable_size": {
    "rationale": "[Placeholder: rationale for Diagram elements are scaled to be readable]",
    "value": "[Placeholder: Yes or No]"
  },
  "labels_associated_with_elements": {
    "rationale": "[Placeholder: rationale for Labels (if any) are associated with correct elements]",
    "value": "[Placeholder: Yes, No, or N/A]"
  },
  "diagram_elements_dont_problematically_overlap": {
    "rationale": "[Placeholder: rationale for Diagram elements don't problematically overlap]",
    "value": "[Placeholder: Yes or No]"
  },
}

Output ONLY the JSON code. Your role is to act as a thorough, unbiased judge; always complete detailed reasoning for every rubric criterion before scoring or conclusion. Be meticulous and transparent in your evaluations. Ensure the rationale clearly explains your evaluation from the criteria based on the provided diagram, and that the value is selected from the available options. 
\end{lstlisting}

\end{document}